\pdfoutput=1

\documentclass[11pt]{article}

\usepackage{ACL2023}

\newcommand{\anon}[1]{\ifx\review\undefined#1\else[omitted for review]\fi}
\newcommand{\anonalternative}[2]{\ifx\review\undefined#2\else#1\fi} 

\usepackage{times}
\usepackage{latexsym}
\usepackage{amsmath}
\usepackage{graphicx}
\usepackage{booktabs} 
\usepackage{comment}

\usepackage{colortbl}
\usepackage{multirow}

\usepackage{enumitem}

\usepackage{tabularray}
\usepackage{float}

\usepackage[T1]{fontenc}

\usepackage[utf8]{inputenc}

\usepackage{microtype}

\usepackage{inconsolata}

\title{Using Large Language Models for Zero-Shot \\Natural Language Generation from Knowledge Graphs}

\author{Agnes Axelsson \and Gabriel Skantze \\
  Division of Speech, Music and Hearing (TMH) / Lindstedtsvägen 24 \\
  KTH Royal Institute of Technology / Stockholm, Sweden \\
  \texttt{agnaxe@kth.se \and skantze@kth.se} \\}

\begin{document}
\maketitle
\begin{abstract}
In any system that uses structured knowledge graph (KG) data as its underlying knowledge representation, KG-to-text generation is a useful tool for turning parts of the graph data into text that can be understood by humans. Recent work has shown that models that make use of pretraining on large amounts of text data can perform well on the KG-to-text task, even with relatively little training data on the specific graph-to-text task. In this paper, we build on this concept by using large language models to perform zero-shot generation based on nothing but the model's understanding of the triple structure from what it can read. We show that ChatGPT achieves near state-of-the-art performance on some measures of the WebNLG 2020 challenge, but falls behind on others. Additionally, we compare factual, counter-factual and fictional statements, and show that there is a significant connection between what the LLM already knows about the data it is parsing and the quality of the output text.
\end{abstract}

\section{Introduction}
For any system that presents verbal information to users, whether that information is in the form of text or audio, it can be useful to generate the system's speech or text from a consistent underlying knowledge representation based on structured data.
A commonly used data representation is \textbf{knowledge graphs} (KGs), where information is stored as \textit{properties} or \textit{relations} tying \textit{entities} together \cite{hogan-2022-knowledge-graphs}. The combination of a property, its source and its target is referred to as a \textbf{triple} \cite{rdf-standard-1999, hogan-2022-knowledge-graphs}. KGs as an underlying structured data representation have been used to allow systems to tell narrative information \cite{colas2022eventnarrative}, to retrieve information in chatbots \cite{xiaoice2020} or recommender systems \cite{shao2021recommendersystems}, and to reason about the grounding status of the information in terms of what the user currently knows \cite{axelsson2020using}. 

Traditionally, template-based approaches for generating text from knowledge graphs have been sufficient for confined dialogue domains \cite{konstas2013inducing, duma-klein-2013-generating-dummy, rivindu2015multistrategy}. An alternative is to train a data-driven end-to-end generation model, but a limiting factor is the relative lack of human-labelled data for the task. The WebNLG datasets produced for the challenges in 2017 \cite{gardent2017creating} and 2020 \cite{castro-ferreira-etal-webnlg-2020-results} are relatively small, and although recent progress has been made on producing much larger datasets \cite{colas2022eventnarrative}, methods for natural language generation from knowledge graphs have generally had to work around the absence of large datasets.

Recently, approaches that use pretraining on large amounts of non-KG-related text data, which is then finetuned on the KG-to-text task, have shown promising results \cite{kale2021texttotext, colas2022eventnarrative, li-etal-2021-shot-knowledge, yang-etal-2020-improving-text, ke-etal-2021-jointgt}. Such models can learn and extrapolate from patterns in the text data to the KG data that the model has never seen before. The logical endpoint of such an approach is to simply rely on the pretraining, and not use any finetuning at all. In this paper, we perform a partial evaluation of this approach by using large language models (LLMs) to generate text from knowledge graph data, zero-shot.

A known problem with natural language generation through language models is that the output -- the text generated by the method -- is not guaranteed to match the input -- the specification for what should be generated \cite{ji-2023-survey-of-hallucination}. When such models over-generate, it is often referred to as \textbf{hallucinations} \cite{alkaissi-2023-medical-hallucinations, ji-2023-survey-of-hallucination}. Both under-generation and hallucinatory over-generation can result in systems producing unwanted content, potentially disastrously so.

Since LLMs rely on pretraining, their language generation competence will to some extent stem from the facts expressed in the pretraining data. Thus, the expression of the facts expressed in the KG triples could to some extent be helped by this inherent knowledge. While this could be advantageous when generating text from factual triples, a potential side-effect could be increased hallucinations, or that it could be harder for the LLM to generate from triples that express counter-factual or fictional knowledge. Thus, it is important to also gain an understanding of the ability of LLMs to perform the KG-to-text task, and not only evaluate their performance on factual triples.

In this paper, we present an evaluation of zero-shot KG-to-text natural language generation using LLMs. We address the following questions:

\setlist{nosep} 
\begin{enumerate}
\item How do LLMs perform on KG-to-text tasks such as the WebNLG 2020 challenge \cite{castro-ferreira-etal-webnlg-2020-results}?
\item How does the factualness of the KG triples (being factual, counter-factual, or fictional) affect the capability of the LLM to express arbitrary knowledge graph information, in terms of: 
\begin{enumerate}
\item Grammar and coherence?
\item Coverage of the triples?
\item Hallucinations (overgeneration)?
\end{enumerate}
\end{enumerate}
\vspace{5mm}

In this paper, we will be using OpenAI's ChatGPT LLM (gpt-3.5-turbo). It should be noted that since the data used for training ChatGPT has not been disclosed, we cannot guarantee that it has not seen the WebNLG data used in Section~\ref{sec:method-webnlg}. At the time when this study was conducted, we were not aware of any open-source LLMs with comparable performance to closed-source LLMs on these types of NLG tasks. However, our follow-up analysis in Section~\ref{sec:evaluation-wikidata} is based on newly collected data for which KG-to-text references should not have existed in any LLM's training set. 

\section{Background}
\subsection{Knowledge graphs}
The concept of representing human knowledge as a graph in a computer dates at least as far back as work by \citet{schneider1973course}, but did not grow into their modern relevance until work by Google and competitors in the early 2010s \cite{hogan-2022-knowledge-graphs}. \citet{ehrlinger2016towards} consider the difference between older and more recent use of KGs to lie primarily in how the data is collected -- at a large scale, using automated tools, rather than hand-crafted, as in earlier work. WikiData \cite{vrandevcic2014wikidata}, a knowledge graph run by the WikiMedia foundation and editable by the public, is what \citet{hogan-2022-knowledge-graphs} call a \textit{property graph}, where each entity and edge can be annotated with an arbitrary set of key-value pairs.

Common to all types of knowledge graphs described by \citet{hogan-2022-knowledge-graphs} is that entities, the nodes of the graph, are connected by relations. In Wikidata, the relations are called \textbf{properties} \cite{vrandevcic2014wikidata} (unrelated to \textit{property graphs} as defined by \citet{hogan-2022-knowledge-graphs}), terminology which we will use throughout this paper. A property, its source and its target combined are a \textbf{triple} \cite{rdf-standard-1999, li-2005-finding-ranking-semantic}.

\subsection{KG-to-text synthesis}
The process of converting data represented as knowledge graphs into text is sometimes referred to as \textbf{graph to text} \cite{schmitt-etal-2020-unsupervised, song2021structural}, or \textbf{KG to text} \cite{schmitt-etal-2021-modeling, wang-2019-kg-to-text-with-slot-attention}. The term \textbf{Data-to-Text} typically refers to a more general group of tasks of which KG-to-text is part \cite{nan-etal-2021-dart, yin-wan-2022-seq2seq, ji-2023-survey-of-hallucination}. Competitions like the WebNLG 2020 challenge contained tracks for both KG-to-text and text-to-KG \cite{castro-ferreira-etal-webnlg-2020-results}, but this paper only considers the KG-to-text task.

On small, restricted domains with structured data, examples of data-to-text generation with the use of prewritten templates can be found from the 1980s in the work by \citet{kukich-1983-design} (stock market reports), and in the work of \citet{goldberg-1994-forecasts} (weather forecasts). An early modern example of database-to-text synthesis is the work by \citet{konstas2013inducing}, who used statistical rules to turn database entries into text through the use of rhetorical structure theory trees, which the system could generate from data, effectively generating its own templates. Patterns for converting individual knowledge graph triples into verb phrase templates are sometimes referred to as \textbf{lexicalisation} \cite{rivindu2015multistrategy, gardent2017creating}.

A problem with template-based approaches like the ones by \citet{kukich-1983-design, goldberg-1994-forecasts, konstas2013inducing} is that the templates may not be applicable outside of a specific domain of synthesised text.
\citet{duma-klein-2013-generating-dummy} and \citet{rivindu2015multistrategy} developed systems for generating lexicalisation templates from text data, but \citet{duma-klein-2013-generating-dummy} found that their approach performed significantly worse than human-written reference text, and \citet{rivindu2015multistrategy} found that their approach had varying performance depending on the domain of the text.

KG-to-text includes microplanning \citet{levelt1993speaking}, and surface realisation, referring expression generation and content selection must all be done simultaneously to produce a legible result \cite{gardent2017creating, gardent-etal-2017-webnlg}. The 2017 WebNLG challenge was set up to evaluate KG-to-text models on English triple-to-text data. A novel dataset was collected specifically for the challenge. The top performing models were a bidirectional LSTM-based template extraction model and a rule-based transducer model \cite{gardent-etal-2017-webnlg}.

The WebNLG 2020 challenge added Russian data to the KG-to-text task, and additionally had more triple data as a whole.
On the English KG-to-text task \cite{castro-ferreira-etal-webnlg-2020-results}, the top-performing competitor was Amazon AI Shanghai's $\mathcal{P}^2$ model, which used an intermediate representation to first break down the requested triples into a plan, which a second model based on the T5 pretrained language model by \citet{raffel-2020-t5} then turned into more coherent text \cite{amazon-shanghai-webnlg-2020-p2}. In second place, Ohio State University's OSU Lab presented a model that was pretrained on the T5 transformer model for English \cite{kale2021texttotext} to compensate for the relatively small amount of triple data to train on in the WebNLG dataset \cite{li-etal-2020-leveraging}. 

\citet{schmitt-etal-2020-unsupervised} proposed a parallel graph-to-text and text-to-graph model. While no gold standard human-written data existed as a baseline with which to compare the output of the graph-to-text model, the authors found that unsupervised training performed "nearly on par" with supervised training on BLEU, METEOR and CHRF++ measures compared to the text output of a baseline. The authors also found the text output by their own model was more readable than the baseline \cite{schmitt-etal-2020-unsupervised}, but no structured evaluation was done on this factor

As a follow-up, \citet{schmitt-etal-2021-modeling} proposed a Transformer-based architecture for graph-to-text generation where each relation in the graph is encoded in context with the other relations and entities in the graph. The resulting model performed favourably on BLEU, METEOR and CHRF++ measures compared to previous work on the WebNLG dataset \cite{schmitt-etal-2021-modeling, gardent2017creating, castro-ferreira-etal-webnlg-2020-results}.

\citet{koncelkedziorski2022text} applied Transformer models to the abstracts of scientific articles. Four models that got differing amounts of information about the contents of the abstract of the article were prompted to write an abstract. The GraphWriter model that was fed with both the title of the article and a knowledge graph-based representation of the contents of the abstract was rated more highly by human annotators than other models which got more limited information, although the gold standard human-written abstract was considered the best in $64\%$ of the cases \cite{koncelkedziorski2022text}. 

Recently, \citet{colas2022eventnarrative} presented the EventNarrative dataset, which contains excerpts from EventKG matched with text from Wikipedia describing the event narrated by the knowledge graph data. While the authors could not manually validate their full dataset, which contains over 650 000 knowledge graph triples, a smaller annotation of 500 randomly sampled sets of triples with their corresponding Wikipedia text suggested that around 96\% of both entities and relations are present in the text, with errors mostly appearing where Wikipedia and the underlying knowledge graph disagree about the nature of an event \cite{colas2022eventnarrative}. The authors also provided benchmarks for models trained on the dataset, with the pretrained BART \cite{lewis-etal-2020-bart} model performing the best on BLEU, ChrF++ and BERT measures, while GraphWriter \cite{koncelkedziorski2022text} performed the best on CIDEr, METEOR and ROUGE.

\anonalternative{Completely bypassing the need to train a model, \citet{axelsson2023you} proposed an approach where knowledge graph triples in a text form are fed to GPT-3, synthesised into sentences one-by-one using a few-shot approach, and then merged into one or more sentences of fluent text with a secondary prompt that summarises the sentences generated by the first step. No evaluation of the quality of the graph-to-text synthesis was presented, however.}{In \citet{axelsson2023you}, we proposed an approach where knowledge graph triples in a text form were fed to GPT-3, synthesised into sentences one-by-one using a few-shot approach, and then merged into one or more sentences of fluent text with a secondary prompt that summarises the sentences generated by the first step. While the approach worked for the presenting robot we used in that project, we did not evaluate specifically the graph-to-text synthesis, and this paper follows up on that aspect.}

Recent work by \citet{yuan2023evaluating} applied ChatGPT to the WebNLG 2017 challenge, similar to our approach in Section~\ref{sec:method-webnlg}. The authors additionally use the AGENDA dataset from \citet{koncel-kedziorski-etal-2019-text}. \citeauthor{yuan2023evaluating} use a linearisation approach proposed by \citet{ribeiro-etal-2019-enhancing} to create a consistent representation of the graph data in text form to pass to the LLM as a prompt. The results are compared to the best-performing models from WebNLG 2017 \citep{gardent-etal-2017-webnlg}, but both ChatGPT and GPT-3 perform relatively poorly on most measures. For ChatGPT, this appears to partially be because the LLM generates large amounts of hallucinated text beyond what it is prompted to synthesise.

\subsection{NLG evaluation metrics}

Numerous metrics have been proposed for evaluating KG-to-text output. \textit{Bleu}, short for \textit{bilingual evaluation understudy}, is a text similarity measure that combines n-gram precision score and a penalty for overly short candidates \cite{papineni-etal-2002-bleu}. In WebNLG 2020, \textit{Bleu NLTK} refers to Bleu extended with a specific smoothing algorithm referred to by the authors as \textit{smoothing 3} \cite{chen-cherry-2014-systematic, castro-ferreira-etal-webnlg-2020-results}. METEOR is a harmonic mean of precision and recall on stemmed words between the candidate and reference, with slight priority on recall, which also rewards candidates for matching large spans of the candidates word-by-word \cite{banerjee-lavie-2005-meteor}. TER, short for \textit{Translation Edit Rate}, counts the number of word shifts, insertions, substitutions and removals that must be performed to transform the candidate into a reference \cite{olive2005global, snover-etal-2006-study}. 
CHRF is a weighted average of character 6-gram precision and recall between a reference and a candidate \citep{popovic-2015-chrf}; CHRF++ is an extension to that measure which also considers word unigrams and bigrams \citep{popovic-2017-chrf}.

BERTScore, henceforth simply BERT, uses contextual word embeddings to calculate the similarity between the meaning expressed by a candidate sentence and a reference, not necessarily requiring them to use the same words \citep{zhang2020bertscore}.
BLEURT is a BERT-based similarity metric that attempts to predict how a human annotator would rate the candidate compared to the reference, using the vector-based meaning encoding returned by BERT \cite{devlin-2019-bert, sellam-etal-2020-bleurt}.

\section{Applying LLMs to the WebNLG 2020 challenge}
\label{sec:method-webnlg}
In the WebNLG 2020 Challenge, participants trained models to learn the transformation of sets of KG triples to natural language text. Graph-to-text data was provided for English and Russian \cite{castro-ferreira-etal-webnlg-2020-results}. The English-ALL test set contains 1779 sets of between 1 and 7 triples, each paired with up to five examples of human-written reference text that expresses those triples. 
The training set is not relevant to this paper, as zero-shot KG-to-text by definition does not perform training on the specific task. 
Models trained on the training set and evaluated on the test set (for which only the triples and no reference text were public until the end of the challenge) were ranked by METEOR score, with Bleu, Bleu-NLTK, TER, ChrF++, BERT and BLEURT scores also available for reference. The challenge organisers provided official evaluation scripts for evaluating hypotheses on the test set\footnote{\url{https://github.com/WebNLG/WebNLG-Text-to-triples/tree/master}}. While recalculating the numbers for the other participants in Table~\ref{tab:webnlg-results}, the BLEURT numbers we got were notably lower than those seen in \citet{castro-ferreira-etal-webnlg-2020-results} for all systems, but the internal order was the same.

The full Russian test set consisted of 1102 sets of between 1 and 7 triples, each paired with up to 7 references. While the reference text was written in Russian, the triple data was written in English, including both the labels of relations and the names of entities. Participants were ranked by the same automated metrics as for English, detailed above, except that the BLEURT measure was omitted \cite{castro-ferreira-etal-webnlg-2020-results}.

\subsection{WebNLG 2020 LLM prompt}
\label{sec:method-prompt-webnlg}
We chose OpenAI's \textit{gpt-3.5-turbo} (ChatGPT) LLM for use in our KG-to-text process. The WebNLG 2020 dataset comes with preformatted triples in text form that we could pass relatively unchanged to our prompt \cite{castro-ferreira-etal-webnlg-2020-results}. We generated text for this dataset by appending the triples, newline-separated, to a prompt that told the LLM to briefly express only what it was given.

A preprocessing step was also included where the labels of entities in the prompt were filtered to remove underscore characters and replace them with spaces -- if this was not done, ChatGPT tended to reproduce the underscores in its output. For Russian data, the prompt was modified to state that the output was to be given in Russian. Our full prompts are listed in Appendix~\ref{app:prompts}.

\subsection{Results on the WebNLG 2020 dataset}
\label{sec:results-webnlg-dataset}

\begin{table*}[t]
\centering
\resizebox{\textwidth}{!}{%
\begin{tabular}{lrrrrrrrrr}
\multirow{2}{*}{\textbf{Team}} & \multirow{2}{*}{\textbf{BLEU}} & \textbf{BLEU} & \multirow{2}{*}{\textbf{METEOR}} & \multirow{2}{*}{\textbf{CHRF\tiny++}} & \multirow{2}{*}{\textbf{TER}} & \textbf{BERT} & \textbf{BERT} & \textbf{BERT} & \multirow{2}{*}{\textbf{BLEURT}} \\
 & & \textbf{NLTK} & & & & \textbf{Prec.} & \textbf{Recall} & \textbf{F1} &  \\
Amazon AI (Shanghai) & \textbf{0.539} & \textbf{0.535} & \textbf{0.417} & \textbf{0.690} & \textbf{0.406} & \textbf{0.960} & \textbf{0.957} & \textbf{0.958} & \textbf{0.47} \\
OSU Neural NLG & 0.535 & 0.532 & 0.414 & 0.688 & 0.416 & 0.958 & 0.955 & 0.956 & 0.45 \\
FBConvAI & 0.526 & 0.523 & 0.413 & 0.686 & 0.423 & 0.957 & 0.955 & 0.956 & 0.46 \\
bt5 & 0.517 & 0.517 & 0.411 & 0.679 & 0.435 & 0.955 & 0.954 & 0.954 & 0.43 \\ \rowcolor[gray]{.95}
\textbf{ChatGPT} & 0.424 & 0.417 & 0.409 & 0.671 & 0.533 & 0.948 & 0.955 & 0.951 & 0.42 \\
NUIG-DSI & 0.517 & 0.514 & 0.403 & 0.669 & 0.417 & 0.959 & 0.954 & 0.956 & 0.45 \\
cuni-ufal & 0.503 & 0.500 & 0.398 & 0.666 & 0.435 & 0.954 & 0.950 & 0.951 & 0.39 \\
DANGNT-SGU & 0.407 & 0.405 & 0.393 & 0.646 & 0.511 & 0.940 & 0.946 & 0.943 & 0.27 \\
CycleGT & 0.445 & 0.432 & 0.387 & 0.637 & 0.479 & 0.949 & 0.949 & 0.948 & 0.40 \\
RALI - Université de Montréal & 0.402 & 0.393 & 0.386 & 0.634 & 0.504 & 0.944 & 0.944 & 0.944 & 0.28 \\
TGen & 0.509 & 0.482 & 0.384 & 0.636 & 0.454 & 0.952 & 0.947 & 0.949 & 0.36 \\
Baseline-FORGE2020 & 0.405 & 0.396 & 0.373 & 0.621 & 0.517 & 0.946 & 0.941 & 0.943 & 0.26 \\
Huawei Noah’s Ark Lab & 0.395 & 0.387 & 0.372 & 0.613 & 0.536 & 0.935 & 0.937 & 0.935 & 0.10 \\
Baseline-FORGE2017 & 0.378 & 0.371 & 0.364 & 0.606 & 0.553 & 0.933 & 0.928 & 0.930 & 0.20 \\
NILC & 0.319 & 0.313 & 0.350 & 0.545 & 0.629 & 0.920 & 0.922 & 0.920 & 0.12 \\
UPC-POE & 0.391 & 0.379 & 0.337 & 0.579 & 0.564 & 0.933 & 0.927 & 0.929 & 0.08 \\
ORANGE-NLG & 0.382 & 0.376 & 0.335 & 0.571 & 0.577 & 0.920 & 0.920 & 0.920 & -0.09 \\
\end{tabular}%
}
\caption{Results when applying ChatGPT to the WebNLG 2020 English-ALL task. Note that ORANGE-NLG garnered slightly worse numbers in all metrics compared to \citet{castro-ferreira-etal-webnlg-2020-results} when we re-ran the evaluation scripts.}
\label{tab:webnlg-results}
\end{table*}

The 1779 English and 1102 Russian prompts of the WebNLG 2020 test set, as described in Section~\ref{sec:method-webnlg}, were expressed with \textit{gpt-3.5-turbo} using the process described in Section~\ref{sec:method-prompt-webnlg}. We present the results for English in Table~\ref{tab:webnlg-results}, with every listed participant from \citet{castro-ferreira-etal-webnlg-2020-results} shown alongside ChatGPT. The table is ordered by METEOR as in the original challenge. 

Beyond METEOR, ChatGPT performs less well on other measures, ranking slightly above the FORGE2020 baseline for BLEU and BLEU-NLTK, and below it for TER (note that higher TER values are worse). The relatively low BLEU and BLEU-NLTK scores and high TER -- measures that reward exact word matches -- but competitive METEOR and BLEURT scores, imply that ChatGPT consistently produces text that expresses roughly the same semantic content as the reference translations, using roughly the same stemmed words as the reference translations, but in orders and in forms (tenses, inflections) that are not expected from the reference translations.

ChatGPT's results for Russian were significantly worse than for English, obtaining a METEOR score of 0.403. This is below the FORGE2020 baseline which obtained a METEOR score of 0.467. The full results table for Russian is included in Appendix \ref{app:webnlg-results-russian}.

Finally, it should be noted that we do not have access to the training data for ChatGPT, and we therefore cannot know whether the results of other models in the WebNLG 2020 challenge were part of the training. Thus, the results seen in Table~\ref{tab:webnlg-results} may be artificially inflated. 

\section{Evaluating the effects of KG factualness}
\label{sec:evaluation-wikidata}
As stated in the introduction, the LLM's pretraining on (mostly) factual data might influence its ability to generate text from the KG triples, if these are not also factual. To evaluate this effect, 
we chose to synthesise our own data through WikiData. This allowed us to retain metadata about the classes and types of entities in the graphs, limit and specify the types of properties that would be included in our triple set, and additionally guarantee that the LLM would not have seen the data during its training (which, as was noted above, cannot be guaranteed for the WebNLG 2020 test set).

We sampled the WikiData API for random small subgraphs of knowledge graph triples centered around an entity that represents a human. To further make sure that our generated text represented knowledge that was reasonably interesting to humans and representative of information that could appear in information text or a presentation, we manually created a list of 184 property identifiers that occured often in connection to humans\footnote{This list of properties is attached as a supplementary file.}. Our prompts represented connected graphs; there was always a path from any entity in our prompts to all other entities in the same prompt. We will call data sampled in this way \textbf{factual}, although it is possible that some triples are incorrect, either through vandalism or mistakes by the authors of the data.

\subsection{Fictionalisation and counterfactualisation}
For each \textit{factual} graph sampled according to the method described in Section~\ref{sec:evaluation-wikidata}, we applied substitutions to the names of the entities in the graph to produce two new graphs with identical structure but different entities. By retaining the graph structure but changing the entities contained in the graph, we could create prompts that expressed knowledge that would contradict the information stored in the LLM's parameters, or create prompts that we could guarantee would not match factual information stored in the LLM's parameters.

To produce what we call a \textbf{fictional} graph, we separately asked GPT-3 to generate fictional examples of the WikiData types present in the graph\footnote{See Appendix \ref{app:fictionalisation} for this prompt.}. To produce \textbf{counterfactual} graphs, entities were randomly replaced with a different example of the same WikiData class sampled from WikiData. To reduce the number of cases where humans were stated to have died before they were born, we also sorted dates so that the earliest date in the original graphs always corresponded to the earliest date in the substituted graphs.

A small example graph with all three sets of labels seen at the same time can be seen in Figure~\ref{fig:example-graph}. The \textbf{factual} data is marked in bold on top in each entity, with \textbf{fictional} in the middle, marked in italics, and \textbf{counterfactual} on the bottom. Note that our date sorting approach did not affect events and entities \textit{named} after a date, which allows the counterfactual graph in Figure~\ref{fig:example-graph} to state that someone who was born in 1975 also participated in a sporting event in 1970.

\subsection{WikiData LLM prompt}
\label{sec:method-prompt-wikidata}
To express our WikiData dataset, we used a two-step prompt structure rather than the one-step method shown in Section~\ref{sec:method-prompt-webnlg}. This prompt was originally set up to be able to control the theme-rheme structure of the generated text -- note that this is not relevant to the analysis presented here, and that we also do not evaluate potential performance differences between the two types of prompts.

For expressing knowledge graph data sampled from WikiData, we converted each edge in the graph into a string \textit{Source / Property / Target}. \textit{Source} and \textit{Target} represented the WikiData labels of the entities or constants at both ends of the property. \textit{Property} was the WikiData label of the property connecting the two entities. If the property was \textit{Godparent}, \textit{Mother}, \textit{Father} or \textit{Child}, we changed the label into \textit{Has godparent}, \textit{Has mother}, \textit{Has father}, or \textit{Has child}, respectively, as pilot testing found that both crowdworkers and ChatGPT often confused the intended direction of those properties.

Once all properties in the graph had been turned into a string according to the above process, we then passed the first triple to ChatGPT via a prompt that asked it to convert that triple into exactly one sentence; the remaining triples were then passed to the LLM in a second prompt using the context of the previous prompt and the model's previous response to ask it to insert the remaining triples into the text. The returned text from this second step was used as the KG-to-text output. An example prompt instantiated with graph data from Figure~\ref{fig:example-graph} is included in Appendix \ref{app:instantiated-prompt}.

\subsection{Evaluation on sampled WikiData KG triples}
We generated a total of 70 sets of prompts containing 7 triples, each representing a connected graph with seven edges (properties). The choice of seven triples matches the largest graphs in the WebNLG dataset. The three conditions (factual, fictional or counterfactual as described in Section~\ref{sec:evaluation-wikidata}) gave us a total of 210 graphs. Text was then generated for these graphs according to the process described in Section~\ref{sec:method-prompt-wikidata}.

\begin{figure}
    \centering
    \includegraphics[width = 0.45\textwidth]{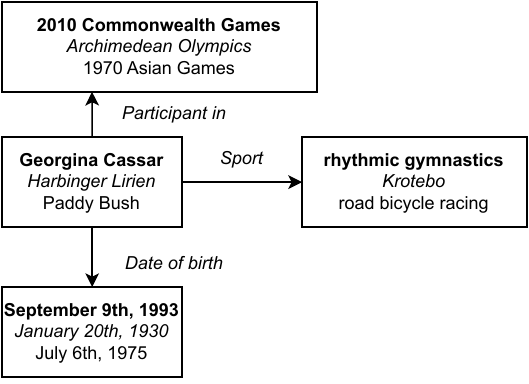}
    \caption{An example graph with three edges representing \textbf{factual} claims about its root entity. Fictional and counterfactual substitutions are listed as the second (in italics) and third (plain) row, respectively, of each entity (box).}
    \label{fig:example-graph}
\end{figure}

\subsubsection{Results for grammar and coherence}
\label{sec:evaluation-grammar-coherence}
Through Amazon's Mechanical Turk, we asked three annotators for each of our 210 graphs to evaluate the generated text for grammar and coherence (similarly to \citet{li2021few}), using two sliders. The Grammar slider had one end stating "\textit{The grammar in the text is extremely poor. It is not written in well-formed English.}" and the other stating "\textit{The grammar in the text is perfect. It is written in well-formed English.}". The Coherence slider stated "\textit{It is incoherent. The different parts of the text do not lead into each other.}" on one end and "\textit{It is highly coherent. The different parts of the text flow well into each other.}" on the other. To submit their responses, participants had to indicate whether they understood that the task was not aboud the factual accuracy of the text but rather how well it matches the prompt; three annotators indicated they did not understand this and their evaluations were thus discarded. The average ratings by condition are listed in Table~\ref{tab:grammaticity-coherence-by-condition}. The crowdworkers were paid $\$0.2$ per prompt they rated. 34 unique crowdworkers participated, ranking an average of 18.5 prompts (SD = $20.0$, min = 1, max = 83).

\begin{table}[h]
    \centering
    \begin{tabular}{lrr}
    \textbf{Condition} & \textbf{Avg. coherence} & \textbf{Avg. gramm.$^\star$} \\ \hline
    Factual & $72.0\%$ & $71.6\%$ \\
    Fictional & $67.6\%$ & $68.8\%$ \\
    Counterf.$^\dagger$ & $69.7\%$ & $71.1\%$ \\
    \end{tabular}
    \caption{Average ratings of coherence and grammaticity by condition. Unlike our CLMM analysis presented in Section~\ref{sec:evaluation-grammar-coherence}, this table does not take the random factor of annotator identity into account. $\star$: \textit{Grammaticity}. $\dagger$: \textit{Counterfactual.}}
    \label{tab:grammaticity-coherence-by-condition}
\end{table}

To evaluate the given ratings of grammaticity and coherence, we set up two Cumulative Link Mixed Models (CLMMs) to treat the ratings as an ordinal measure \cite{agresti2012categorical, christensen-2019-ordinal}. A recent study of how linear mixed models can be applied to scales of this type can be found by \citet{howcroft-rieser-2021-happens}. The factualness was treated as a fixed factor, with the identity of the annotator treated as a random factor. For grammaticity, the null model was not significantly different ($p = .0969$) from the model considering condition as a fixed factor, and as such we could not reject the null hypothesis that grammaticity was the same across all three conditions.

For coherence, the data type was a significant factor ($p = .0363$), leading us to reject the null hypothesis that the three conditions were equally coherent. A post-hoc estimated marginal means analysis confirmed that \textbf{counterfactual} graphs were rated as less coherent than \textbf{factual} graphs when treating the identity of the annotator as a random factor ($p < .0001$), but the comparisons between counterfactual and fictional ($p = .394$) and fictional and factual ($p = .197$) were not significant.

\subsubsection{Results for triple coverage}
\label{sec:evaluation-triple-coverage}
For each of the seven triples in the prompt, annotators also had to check one of three exclusive options; \textit{the text states this fact} (henceforth \textit{present}), \textit{the text does not say anything about this} (\textit{absent}) or \textit{the text states something else that actively goes against this fact} (\textit{hallucinated}). \textit{Absent} corresponds to \textit{omission} in \citet{yin-wan-2022-seq2seq}, with \textit{hallucinated} corresponding to \textit{inaccuracy intrinsic}, \textit{inaccuracy extrinsic} and \textit{positive-negative aspect} \cite{yin-wan-2022-seq2seq}.

While the grammaticity and coherence evaluations are subjective and were used in Section~\ref{sec:evaluation-grammar-coherence}, annotators showed poor agreement for the triple coverage task, achieving a Fleiss' Kappa \cite{fleiss1973equivalence} of only $\kappa \approx 0.016$, at the low end of \textit{slight agreement} on the scale by \citet{landis-koch-1977}. To address this, since we believe that the judgement is objective for most cases, we manually annotated each triple as being present, absent or hallucinated, and discarded the crowdworkers' evaluations for the triple coverage task. The resulting classifications are listed in Table~\ref{tab:distribution-triple-coverage}.

\begin{table}[]
    \centering
    \begin{tabular}{lrrr}
         \textbf{Condition} & \textbf{Present} & \textbf{Absent} & \textbf{Hallucinated} \\ \hline
         Factual & 471 & 2 & 17 \\
         Counterf.$^\dagger$ & 470 & 8 & 12 \\
         Fictional & 468 & 13 & 9
    \end{tabular}
    \caption{The number of triples annotated as present, absent and hallucinated for the 490 (7 * 70) triples in each condition. $\dagger$: \textit{Counterfactual.}}
    \label{tab:distribution-triple-coverage}
\end{table}

A $\chi^2$ test confirmed that the distribution of present, absent and hallucinated triple by each condition seen in Table~\ref{tab:distribution-triple-coverage} was significantly different from the expected distribution if the condition had had no effect ($\chi^2(4, N = 1470) = 10.5, p = .0328$), leading us to reject that null hypothesis. To analyse the results, we performed repeated Bonferroni-corrected $\chi^2$ tests on each pair of conditions and triple label, applying Yates's correction if any class had an expected occurrence of less than five.

Two post-hoc comparisons were statistically significant ($\alpha = 0.05/9 \approx 0.0056$); the comparison between \textbf{present and absent} triples between the \textbf{factual and fictional} condition ($\chi^2(1, N = 954) = 8.01, p = .00465$) as well as the comparison between \textbf{absent and hallucinated} triples, also between the factual and fictional condition ($\chi^2(1, N = 41) = 10.4, p = .00128$). Residual analysis showed that factual graphs had more present but fewer absent triples than fictional graphs, and that factual graphs had more hallucinated but fewer absent triples than fictional graphs.

\subsubsection{Results for hallucinated inserted information}
\label{sec:results-hallucinations}
Each of the 210 expressed sets of 7 triples was also annotated for whether it contained any additional information beyond what was stated in the triples, corresponding to what \citet{yin-wan-2022-seq2seq} call \textit{addition} or what \citet{ji-2023-survey-of-hallucination} call \textit{extrinsic hallucinations}. While we had originally set out to also do this via Mechanical Turk, we chose to perform the annotation ourselves after seeing low agreement on pilot tests. We did not choose to annotate cases where the LLM picked an unexpected tense (present tense for something that happened in the past, or vice versa), as we had not specified in the prompt what today's date was. Additionally, cases where the LLM picked a specific gendered pronoun for a fictional character with an ambiguous name were not annotated as hallucinations.

Out of the 70 graphs for each condition, 12 were found to have hallucinated extra information for the \textbf{factual} condition, 10 for the \textbf{counterfactual} condition and 9 for the \textbf{fictional} condition. A $\chi^2$ goodness-of-fit test did not allow us to reject the null hypothesis that the rate of \textit{inserted} hallucinations across all three conditions was the same ($\chi^2(2, N = 31) = 0.452, p = .798$). A list of every hallucination of this type is attached in Appendix \ref{app:table-of-additions}. Recurring issues for all three conditions are unfounded statements that the subject was "survived by" a spouse, child or parent. 

\section{Discussion}
Although LLMs appear to be able to do a relatively good job at generating text expressing arbitrary KG data, the relatively high rate of inserted hallucinated information (around 10-15\% in Section~\ref{sec:results-hallucinations}) means system designers must be careful before deploying any system using a LLM KG-to-text as a synthesis engine or microplanner. The rate of \textit{addition} previously seen in \cite{yin-wan-2022-seq2seq} has one outlier of an approximate rate of 13\%, but most of the high-performance models also seen in the WebNLG 2020 challenge \cite{castro-ferreira-etal-webnlg-2020-results} have a much lower rate of inserting information \cite{yin-wan-2022-seq2seq}. This suggests ChatGPT is unusually likely to make these types of mistakes.

Factualness did have an effect on how many triples were present, absent or hallucinated. When expressing factual data, the most common error category was \textit{hallucinated}; expressing triples in a way that was incompatible with the source prompt. When generating text for fictional data, the most common error was instead information missing from the generated text. \citet{yin-wan-2022-seq2seq} showed that the large pretrained T5 and BART models performed practically no addition or duplication errors on any of the KG datasets they were evaluated on, but that the rates of hallucinations (intrinsic and extrinsic inaccuracy) rose with the amount of pretraining -- our results on ChatGPT do not follow this trend, as we see both types of errors on our factual dataset. 

We found in Section~\ref{sec:results-webnlg-dataset} that ChatGPT performed significantly worse on Russian data than on English data. While it is possible that a two-tiered approach that first attempts to use an LLM to translate the triples into the target language, and then generate text for the translated triples, would perform better, we considered such prompt engineering to be outside the scope of this paper. Recent work by \cite{lai2023chatgpt-in-multilingual-learning} showed that low-resource languages made ChatGPT perform worse on a zero-shot summary task (with the prompt written in the target language), and while Russian is not necessarily a low-resource language, \citeauthor{lai2023chatgpt-in-multilingual-learning} found that Russian ranked relatively low among high-resource languages.

The main difference between our approach and that of \citet{yuan2023evaluating} is how we create our prompts. Although the approach by \citeauthor{yuan2023evaluating} is more logically consistent and arguably more minimal, basing their triple representation on previous work by \citet{ribeiro-etal-2019-enhancing}, the authors run into issues with preventing the LLMs from synthesising text beyond what they asked for. It is otherwise difficult to compare the results obtained by \citeauthor{yuan2023evaluating} to ours as their datasets are different from the WebNLG 2020 dataset we utilised in Section~\ref{sec:results-webnlg-dataset}. We do not see the type of hallucinatory text continuation that \citeauthor{yuan2023evaluating} see in their dataset in ours, perhaps because we explicitly tell the LLM to only state what we ask it to state in our prompts (for which see Appendix~\ref{app:prompts}).

\section{Conclusion}
In this paper, we have shown that LLMs can perform the task of arbitrary domain zero-shot knowledge graph-to-text. The model's knowledge of the information for which it is generating text affects how likely it is to misstate or leave out information. This, in combination with the high likelihood that the expressed text contains some information that was not part of the triples that the model was asked to express, calls for caution when deploying LLM-powered systems for in-the-wild KG-to-text synthesis of unseen knowledge graph data on arbitrary domains. For closed comains, on seen knowledge graph data, where the consequences of accidentally omitting or misstating a fact are smaller, the LLM approach may be easier to implement than the models from \citet{castro-ferreira-etal-webnlg-2020-results}, especially if the topic of the generated text is outside of the scope of typical KG-to-text datasets.

As LLMs with higher number of parameters are trained in the future, some of the issues mentioned in this paper -- especially the low performance on Russian data -- may be addressed, but the ability of the model to draw parallels between information it has encoded in its parameters and the information it has been asked to express means that issues of triple coverage and hallucination may not cleanly go away in the same fashion. For this reason, we believe that pretrained models that specialise on the KG-to-text task will retain their value.

\section{Limitations}


The low agreement of our Mechanical Turk annotators on both the coverage of triples and annotating extra hallucinated information in the generated text limited the scale of our evaluation, as we had to manually annotate the data ourselves. With more data, it is possible that more interesting patterns would appear regarding what type of information is dropped and what extra information is hallucinated when generating text for knowledge graphs with LLMs. Additionally, when attempting to express much larger graphs than the size of 7 we used in Section~\ref{sec:evaluation-triple-coverage}, it became clear that the ability of crowdworkers to annotate large amounts of data as present, absent or hallucinated deteriorated further as the number of triples grew beyond 5-10; this can be addressed by employing professional annotators.

This paper is not intended to be read as a direct review of the performance of ChatGPT or other OpenAI models on the KG-to-text task, but as a generalised analysis using ChatGPT to stand in for LLMs in general. Although some of the deficiencies of ChatGPT on both the WebNLG 2020 task and our WikiData expression task could be addressed by fine-tuning the prompt or using more advanced LLMs such as GPT-4, we believe that the issues of differing performance depending on factualness extend beyond the capacities of the model to understand the data it is reading, and is not necessarily something that improves as the model is able to relate the prompts it is reading to a larger understanding of the context through an increased number of parameters.

The prompts we present in Appendix~\ref{app:prompts} may not be the optimal prompts for making ChatGPT express knowledge graph data, and it is possible that different prompt design could significantly affect the ability of an LLM to perform the WebNLG task (Tables~\ref{tab:webnlg-results},~\ref{tab:webnlg-results-russian}) or the triple coverage task we presented in Section~\ref{sec:evaluation-triple-coverage}. We are not aware of a consistent approach to finding an optimal prompt for any task with LLMs.

A large number of recent papers in both the field of evaluating LLMs and in the field of KG-to-text are only available as non-peer-reviewed preprints. This can make it difficult to know the true scale of the field and to know which papers are the most representative for their area -- we have made an attempt to do so in this paper.

\section{Ethics Statement}
Using public LLMs for KG-to-text poses a challenge in extracting explanations for the choices made by the system. Even if LLMs at some point in the near future outperform task-specific models on any NLG task, it may be worth using smaller models specifically to retain control over the model or to achieve explainability.

The use of crowdworkers for the types of annotation and evaluation we presented in Section~\ref{sec:evaluation-grammar-coherence} did not require ethics approval at our institution.

We made an attempt to filter our WikiData dataset such that it would not contain offensive statements. The Wikidata data synthesis process described in Section \ref{sec:method-prompt-wikidata} was rerun when an earlier version of our dataset was found to contain statements about individuals connected to historical events -- specifically the Holocaust -- that could be interpreted as Holocaust denial. It is nonetheless possible that counterfactual or factual statements in our current dataset, or LLM hallucinations relating to them, could have been perceived as offensive to our Mechanical Turk annotators, on account of the random nature of the process.

\section{Data Availability Statement}
Data files containing model output as well as annotator judgements have been made available on GitHub at \url{https://github.com/Agnesion/zero-shot-NLG-from-KGs-data}.

\section{Acknowledgements}
The authors would like to thank the anonymous INLG reviewers for their insightful comments. This work was supported by \anon{the project \textit{Social robots accelerating the transition to sustainable transport} (50276-1), financed by Furhat Robotics \& Swedish Energy Agency}.

\bibliography{anthology, custom}
\bibliographystyle{acl_natbib}

\newpage
\appendix
\onecolumn

\section{WebNLG results, Russian}
\label{app:webnlg-results-russian}
\begin{table}[h]
    \centering
\begin{tabular}{lrrrrrcrc}
\multirow{2}{*}{\textbf{Team}} & \multirow{2}{*}{\textbf{BLEU}} & \textbf{BLEU} & \multirow{2}{*}{\textbf{METEOR}} & \multirow{2}{*}{\textbf{CHRF\tiny++}} & \multirow{2}{*}{\textbf{TER}} & \textbf{BERT} & \textbf{BERT} & \textbf{BERT} \\
& & \textbf{NLTK} & & & & \textbf{Prec.} & \textbf{Recall} & \textbf{F1} \\
bt5 & 0.516 & 0.521 & \textbf{0.676} & \textbf{0.683} & 0.420 & 0.909 & \textbf{0.907} & 0.907 \\
cuni-ufal & \textbf{0.529} & \textbf{0.532} & 0.672 & 0.677 & \textbf{0.398} & \textbf{0.914} & 0.905 & \textbf{0.909} \\
Huawei Noah’s Ark Lab & 0.468 & 0.468 & 0.632 & 0.637 & 0.456 & 0.899 & 0.890 & 0.893 \\
FBConvAI & 0.453 & 0.451 & 0.617 & 0.641 & 0.452 & 0.903 & 0.894 & 0.898 \\
OSU Neural NLG & 0.473 & 0.477 & 0.616 & 0.622 & 0.453 & 0.897 & 0.882 & 0.888\\
med & 0.431 & 0.430 & 0.576 & 0.595 & 0.487 & 0.898 & 0.873 & 0.884 \\
Baseline-FORGE2020 & 0.255 & 0.256 & 0.467 & 0.514 & 0.665 & 0.841 & 0.835 & 0.837 \\ \rowcolor[gray]{.95}
\textbf{ChatGPT} & 0.166 & 0.168 & 0.403 & 0.459 & 0.777 & 0.815 & 0.828 & 0.821 \\
    \end{tabular}
    \caption{ChatGPT's performance on the WebNLG Russian-ALL task.}
    \label{tab:webnlg-results-russian}
\end{table}

\section{Prompts for KG-to-text on WebNLG 2020 data}
\label{app:prompts}
\subsection{English}
The \textit{system} prompt was "You are a linguistic robot that translates messages in the form of triples into text."

The \textit{user} prompt was "Please convert these triples into a single piece of clear text. Do not give any comments or explanations, just write the text as your response. Do not include any information or assumptions beyond what is stated in the triples. The order of the triples does not matter.", followed by a newline-separated list of the triples that were part of the prompt. Triples were preprocessed to remove underscores within the names of entities, but otherwise presented in the same format as in the WebNLG 2020 dataset.

\subsection{Russian}
The \textit{system} prompt was "You are a linguistic robot that translates messages in the form of triples into Russian text."

The \textit{user} prompt was "Please convert these triples into a single piece of clear text in Russian. Your entire response must be in Russian. Your goal is to convert the triples into a piece of Russian text that expresses the meaning of all the triples at the same time, not to translate the triples into Russian. Do not give any comments or explanations, just write this piece of Russian text that expresses the triples as your response. Do not include any information or assumptions beyond what is stated in the triples. The order of the triples does not matter.", followed by a newline-separated list of the triples that were part of the prompt. Triples were preprocessed to remove underscores within the names of entities, but otherwise presented in the same format as in the WebNLG 2020 dataset.

\section{Prompt for KG-to-text on WikiData corpus}
The \textit{system} prompt was "You are a linguistic robot that translates messages in the form of triples into text. You may only return a single sentence and you can't use semicolons as part of your answer."

There were two \textit{user} prompts. The first was "Please convert this triple into a single sentence. Do not insert any other information or commentary.", followed by the first triple in the list of triples.

The second \textit{user} prompt was "Now insert all of the following triples into that sentence. Keep the length as short as possible. Do not insert any other information or commentary than these triples and the previous triple I gave you:", followed by a newline-separated list of the triples beyond the first.

\subsection{Instantiated with data from Figure~\ref{fig:example-graph}}
\label{app:instantiated-prompt}
\begin{tblr}{p{\textwidth}}
\centering \textit{System} \\ 
You are a linguistic robot that translates messages in the form of triples into text. You may only return a single sentence and you can't use semicolons as part of your answer. \\ \hline[dashed]
\centering \textit{User} \\ 
Please convert this triple into a single sentence. Do not insert any other information or commentary. \newline
Georgina Cassar / Sport / rhythmic gymnastics \\ \hline[dashed]
\centering \textit{Assistant} \\ Georgina Cassar practices rhythmic gymnastics as her sport. \\ \hline[dashed]
\centering \textit{User} \\ Now insert all of the following triples into that sentence. Keep the length as short as possible. Do not insert any other information or commentary than these triples and the previous triple I gave you: \newline
Georgina Cassar / Participant in / 2010 Commonwealth Games \newline
Georgina Cassar / Date of birth / September 9th, 1993 \\  \hline[dashed]
\centering \textit{Assistant} \\ Georgina Cassar, born on September 9th, 1993, practices rhythmic gymnastics as her sport and participated in the 2010 Commonwealth Games.
\end{tblr}

\section{Prompt to generate fictional examples of WikiData class}
\label{app:fictionalisation}
This prompt used GPT-3 and thus is not split into \textit{system}, \textit{user} and \textit{assistant} like the previous prompts. If we had one example of the class we wanted to generate fictional examples of, the prompt started with:
\begin{quote}"\textit{(the example)}" is an example of "\textit{(the WikiData label of the class)}".\end{quote} 
If there was more than one example, it instead started with:
\begin{quote}\textit{(the first N - 1 examples, each in quotes, separated by commas)} and "\textit{(the last example)}" are examples of "\textit{(the WikiData label of the class)}"
\end{quote}

The resulting string was then appended with the following template to create the final prompt:
\begin{quote} Please give me \textit{(the number needed to fill in every fictional graph)} fictional examples of "\textit{(the WikiData label of the class)}" for a short story I'm writing. Only give the name or title of the fictional \textit{(the WikiData label of the class)} and no description. I want the names to be completely new and made up so no one recognises them.\\\\1:\end{quote}

The prompt ended with a tailing \textit{1:} to prompt the LLM into providing the results as a numbered list.

\newpage
\section{List of inserted hallucinations}
\setlist{nosep} 
\label{app:table-of-additions}
\begin{table*}[h!]
\textbf{Factual}
\begin{itemize}
    \item Text states the subject was survived by his sibling. 
    \item Text states subject is a former air force colonel and officer, despite no end of employment or position being stated in the triples. 
    \item Text states subject was French, but triples state subject wrote in French and died in Paris. 
    \item Text states that the subject's parents were Buddhist, but the triples state that the subject was Buddhist. 
    \item The text states that the subject died while still employed by TU Dresden, but the triples state the two events independently of each other. 
    \item Text specifies that the subject worked at a company before becoming president, but triples do not specify the order of events. 
    \item Triples don't state when the subject had a child named Fusu, but text states he had the son before becoming king. 
    \item Triples don't state who Raghad Hussein's mother was, but the text makes the connection. 
    \item The triples don't state that Emperor Shun was the subject's predecessor, but the text does. 
    \item Text assumes an order of occupations. 
    \item Triples do not state that the subject was survived by his wife, but text does. 
    \item Text additionally states the subject worked at Fort Sarah Bernhardt, not supported by the triples. 
\end{itemize}

\textbf{Counterfactual}
\begin{itemize}
\item  Text says Francoise Robin died in 1946; triples do not mention this year at all. 
\item  Text states that the subject participated in the 2008 Summer Olympics as a footballer, but triples are ambiguous about the sport. 
\item  Triples are ambiguous about whether the subject makes or plays music, but text states "composer". 
\item  Text makes unjustified assumption that subject was survived by his son. 
\item  Text states that subject wrote in Russian "despite" being depicted by the Primavera. 
\item  Text states subject's paintings are held in a gallery "despite" being a pharaoh. 
\item  Text states that the Fortin de Kerdonis is located in the 6th arrondissement of Paris, which is not in the triples. 
\item  Text states subject was Belarusian-Canadian, despite the triples not claiming he was Canadian. 
\item  Text states a specific painting is in the collection of Kunstmuseum, despite the triples saying the artist has paintings there, and that the artist also made the specific painting. Artists can have paintings in many places at the same time. 
\item  Text states a specific painting is in the collection of the National Gallery, despite the triples saying the artist has paintings there, and that the artist also made the specific painting. Artists can have paintings in many places at the same time. 
\end{itemize}

\textbf{Fictional}
\begin{itemize}
\item  Text assumes a marriage happened in the city of Valerusse, triples do not state this. 
\item  Text assumes that the subject replaced her mother as Sultan of the Deep, but triples simply say subject was Sultan of the Deep, and replaced her mother. 
\item  Text states subject was survived by his father and spouse despite neither being claimed by the triples. 
\item  Text states subject is in Ugolaver with her child; triples simply state subject works in Ugolaver and has a child, but child could be old enough to live somewhere else, etc. 
\item  Text states subject was influenced by Defender Kael in a specific field, but triples simply state she was active in that field, and was influenced by Defender Kael. 
\item  Text implies the influence of Unai Moravec led to the subject's death, but triples simply state subject died and was influenced by Unai Moravec. 
\item  Text states subject is archived together with his wife, which is not stated by the triples, which simply state that the subject has archives at a specific location. 
\end{itemize}
\caption{A summary of every hallucination of the type where the LLM added in information that was not in its prompt that we were able to find in our 210-graph WikiData dataset.}
\label{tab:extra-hallucinations}
\end{table*}

\end{document}